\documentclass[twocolumn, switch]{article} 

\usepackage{preprint}

\usepackage{amsmath, amsthm, amssymb, amsfonts}

\usepackage[numbers,square]{natbib}
\usepackage[utf8]{inputenc}	
\usepackage[T1]{fontenc}	
\usepackage{xcolor}		
\usepackage[colorlinks = true,
            linkcolor = purple,
            urlcolor  = blue,
            citecolor = cyan,
            anchorcolor = black]{hyperref}	
\usepackage{booktabs} 		
\usepackage{nicefrac}		
\usepackage{microtype}		
\usepackage{lineno}		
\usepackage{float}			

\usepackage{lipsum}		

\usepackage{newfloat}
\DeclareFloatingEnvironment[name={Supplementary Figure}]{suppfigure}
\usepackage{sidecap}
\sidecaptionvpos{figure}{c}

\usepackage{titlesec}
\titlespacing\section{0pt}{12pt plus 3pt minus 3pt}{1pt plus 1pt minus 1pt}
\titlespacing\subsection{0pt}{10pt plus 3pt minus 3pt}{1pt plus 1pt minus 1pt}
\titlespacing\subsubsection{0pt}{8pt plus 3pt minus 3pt}{1pt plus 1pt minus 1pt}

\usepackage{tikz,xcolor,hyperref}

\definecolor{lime}{HTML}{A6CE39}
\DeclareRobustCommand{\orcidicon}{
	\begin{tikzpicture}
	\draw[lime, fill=lime] (0,0)
	circle [radius=0.16]
	node[white] {{\fontfamily{qag}\selectfont \tiny ID}};
	\draw[white, fill=white] (-0.0625,0.095)
	circle [radius=0.007];
	\end{tikzpicture}
	\hspace{-2mm}
}
\foreach \x in {A, ..., Z}{\expandafter\xdef\csname orcid\x\endcsname{\noexpand\href{https://orcid.org/\csname orcidauthor\x\endcsname}
			{\noexpand\orcidicon}}
}

\title{Robot Metacognition: Decision Making with Confidence for Tool Invention}

\usepackage{xwatermark}
\newwatermark[firstpage,color=gray!60,angle=90,scale=0.32, xpos=-4.05in,ypos=0]{\href{https://doi.org/}{\color{gray}{  }}}
\newwatermark[firstpage,color=gray!60,angle=90,scale=0.32, xpos=3.9in,ypos=0]{\href{https://doi.org/}{\color{gray}{     }}}
\newwatermark[firstpage,color=gray!90,angle=0,scale=0.28, xpos=0in,ypos=-5in]{*correspondence: \texttt{ajith.anilmeera@ru.nl}}

\usepackage{authblk}

\author[1\thanks{\tt{ajith.anilmeera@ru.nl}}]{Ajith Anil Meera}
\author[2]{Poppy Collis}
\author[3]{Polina Arbuzova}
\author[4,5]{Abián Torres}
\author[2]{Paul F Kinghorn}
\author[4]{Ricardo Sanz}
\author[5]{Pablo Lanillos}

\affil[1]{Donders Institute, Radboud University, Nijmegen, The Netherlands}
\affil[2]{School of Engineering and Informatics, University of Sussex, Brighton, United Kingdom}
\affil[3]{Adaptive Systems, Department of Computer Science, Humboldt-Universität zu Berlin, Germany}
\affil[4]{Autonomous Systems Laboratory, Universidad Politécnica de Madrid, Spain}
\affil[5]{Neuro AI and Robotics, Cajal Neuroscience Center, Spanish National Research Council, Spain}

\begin{document}

\twocolumn[ 
  \begin{@twocolumnfalse} 

\maketitle

\begin{abstract}
Robots today often miss a key ingredient of truly intelligent behavior: the ability to reflect on their own cognitive processes and decisions. In humans, this self-monitoring or metacognition is crucial for learning, decision making and problem solving. For instance, they can evaluate how confident they are in performing a task, thus regulating their own behavior and allocating proper resources. Taking inspiration from neuroscience, we propose a robot metacognition architecture centered on confidence (a second-order judgment on decisions) and we demonstrate it on the use case of autonomous tool invention. We propose the use of confidence as a metacognitive measure within the robot decision making scheme. Confidence-informed robots can evaluate the reliability of their decisions, improving their robustness during real-world physical deployment. 
This form of robotic metacognition emphasizes embodied action monitoring as a means to achieve better informed decisions. We also highlight potential applications and research directions for robot metacognition.
\end{abstract}
\vspace{0.35cm}

  \end{@twocolumnfalse} 
] 



\section{INTRODUCTION}
Robots are increasingly being deployed in uncertain and dynamic environments, from manipulating fragile objects to inventing new tools. However, they still lack the key ability to judge their decisions and identify when to trust them. This is where machine metacognition comes in: the ability of a robot to monitor its own uncertainty and make risk-aware decisions accordingly. Embedding such awareness could fundamentally redefine autonomy, enabling robots that not only act, but also know when and how to act in a way that reflects their self-estimated reliability. In this work, we consider confidence to be the practical handle on metacognition as the system’s estimate of how reliable its current inference is. Everything that follows embraces confidence as the key quantity that operationalizes metacognitive ability. This interpretation opens a frontier in robotics where confidence becomes the driving force of creative, self-aware problem solving, bringing machines one step closer to the reflective intelligence of the human mind. Unlike in disembodied AI systems, confidence in robotics is uniquely tied to physical action, as decisions directly shape interactions in the real world. Incorporating confidence across all decision levels allows robots to act with awareness of their own uncertainty. For example, during tool selection in real-world tasks with high uncertainty, the robot should avoid risks by selecting a robust tool that balances certainty and performance~\cite{meera2024confidence}. It should select a reliable tool with high confidence, rather than fixating on pure performance. 

In order to ground these ideas, we consider tool invention as a concrete domain where confidence-based metacognition could have a transformative impact. Despite major advances in robotic manipulation, autonomous tool invention remains an unsolved challenge. Modern robots can select and use tools from a predefined set, but lack the capacity to design novel tools for new tasks \cite{liu2023learningdesignusetools}. This limitation hampers their utility in dynamic settings such as construction or disaster response, where adaptive physical intelligence is crucial to navigate unpredictable and unstructured environments. 
Autonomous tool invention provides a natural testbed for metacognition since it requires an agent to reason about its own thinking process. An agent must be able to recognize when existing strategies are insufficient, decide whether to persevere or adapt, and then evaluate whether novel solutions might succeed. Critically, confidence can serve as a feedback signal: high confidence validates current strategies, while persistent low confidence signals the need for adjustment or redesign. Metacognitive robots can use this as a signal to guide decision making and iteratively refine their approaches.

Implementing metacognition in robots, despite its potential, is challenging, primarily because its translation from neuroscientific evidence to mathematical formalization requires strong assumptions and conceptualizations \cite{da2022active}. Secondly, from a pure engineering perspective, it should be tractable and applicable to robotic tasks. In this paper, we take a step towards addressing this challenge by $i)$ summarizing how humans use metacognition for decision making; $ii)$ proposing an architecture for robot metacognition; $iii)$ exploring its applicability in a challenging use case of autonomous tool design, discovery, and invention; and $v)$ outlining further research directions. The core contributing idea of our work is the introduction of a confidence block into the general robot decision-making scheme that we argue would furnish robots with metacognitive capabilities.

\section{HUMAN METACOGNITION FOR DECISION MAKING}
In this section, we explore human metacognition and delineate the mathematical formalisms that attempt to model it.

\subsection{Metacognition in humans} 

Metacognition is the ability to monitor and reflect on one’s own cognitive processes. First introduced in psychology, it has been described as “thinking about thinking” and encompasses both awareness of mental states and the regulation of strategies based on that awareness \cite{Nelson1994, Fleming2024}.
In humans, metacognition is often seen as a key expression of intelligence: it supports adaptive learning \cite{Cortese2022}, enables individuals to recognize and correct errors, guides exploration and action selection \cite{Kepecs2008}, and plays an important role in communication and cooperation by making internal states available for social interaction \cite{Dunstone2018}. 

Functionally, metacognition can be understood as the process by which a system evaluates the reliability of its own operations and adjusts behavior accordingly. A central manifestation of metacognition in humans is confidence. Confidence judgments are second-order (metacognitive) evaluations of the reliability of a decision or an action. Humans naturally experience a sense of confidence in their perceptions and choices and can report it explicitly when asked. This provides a signal that can be compared with objective performance and applied in domains such as perception, memory, decision making, and action \cite{Fleming2024}.

Importantly, researchers distinguish between explicit metacognition -- the conscious, reportable awareness of one’s cognitive states -- and implicit metacognition, which manifests itself in behavioral adjustments without conscious access. While explicit metacognition involves subjective experience and verbal report, implicit metacognition can be observed through performance adaptations such as changes in response time, information seeking, or strategy selection based on task difficulty or uncertainty. This broader definition provides a foundation for the implementation of metacognition in robots, where monitoring and adapting to one’s own limitations are crucial for robust and adaptive decision making \cite{Cortese2022}. This view is particularly relevant for tool use applications, where effective problem solving relies on evaluating ongoing performance and revising strategies when necessary.

\subsection{Mathematical models of confidence}
Modeling confidence mathematically requires both a model of the underlying decision-making process and a model of confidence in that decision. Several models have been developed to explain how humans form and report confidence in their decisions. These include various models based on signal detection theory (SDT), drift diffusion models (DDM), and post-decisional evidence accumulation models just to name a few \cite{Shekhar2024ConfidenceModels}. 

Despite their apparent diversity, most of these metacognitive models share a common computational foundation to describe decision making under uncertainty from noisy sensory data. This is precisely the regime in which Bayesian inference provides the mathematically optimal solution, and as such many of these frameworks can be understood as special cases of Bayesian reasoning \cite{khalvati2021bayesian}. Bayesian inference derives its name from Bayes' rule, which is a mathematical formula for inverting conditional probabilities. It states the rule by which we can express (posterior) probabilities for perceptions based on a model describing the data-generating process. We are also free to introduce utility functions in this framework, allowing us to generalize beyond perception to both control and decision-making scenarios in which we must choose between available actions to get the best expected reward. 

Importantly, this belief-based reasoning naturally has an intuitive definition of confidence as the uncertainty in a given posterior probability. Again, many different functions of the posterior distribution have been ascribed to best capture this uncertainty or spread mathematically as a model of confidence \cite{ma2023chapter3}. Here, we propose that measuring the entropy of posterior probabilities is the most general method that makes minimal assumptions about the form of the distribution.

\begin{figure*} 
\centerline{\includegraphics[scale = 0.75]{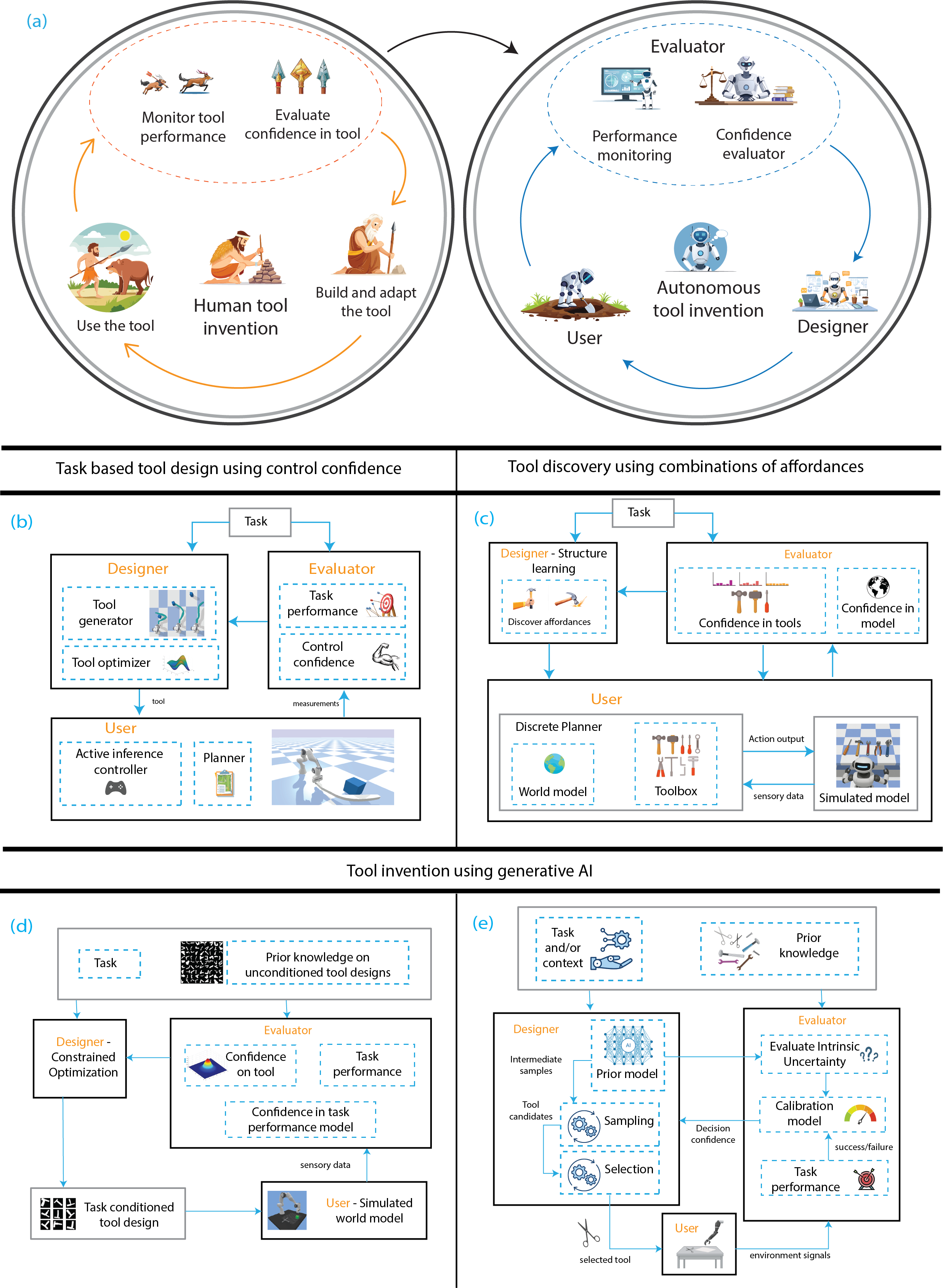}}
\caption{Proposed robot metacognitive architecture narrowed for tool invention. (a) The role of human metacognition during tool invention inspiring our Evaluator-Designer-User scheme. Panels (b–e) then show three levels of tool based applications (tool design, tool discovery and tool invention) all of which are instances of the same Evaluator–Designer–User loop. (b) Tool design using control confidence (from \cite{meera2025designing}) (c) Tool discovery from combination of known affordances (based on \cite{collis2023understanding}) (d-e) Tool invention using generative AI using (d) task-conditioned fine-tuning and (e) decision confidence.}\vspace*{-5pt}
\label{fig:summary}
\end{figure*} 

\section{ROBOT METACOGNITION}
Inspired by human metacognition, we conceptualize robot metacognition for challenging robot problems. In continuous-time dynamic problems, such as controlling a robot arm or stabilizing a drone, this metacognitive ability can be implemented through control confidence, which reflects how well the robot believes that it can keep things under control despite noise or uncertainty. If that confidence drops, it may need to adjust its strategy. In discrete decision-making problems, such as choosing a tool or selecting the next action in a plan, confidence works more like a gut check. Is this choice reliable, given what the robot knows? Low confidence here might lead the robot to explore other tools, gather more information, or switch to a safer backup plan. By treating confidence as a common thread across both continuous and discrete domains, metacognition lets robots reflect, adapt, act, and extend their abilities through informative decisions.



\begin{table*}[h]
 \caption{Types of confidences and their definitions.} \label{tab:conf}
  \centering
  \begin{tabular}{p{4cm}p{4cm}p{7cm}}
    \toprule
    Confidence type& 
Mathematical formulation& 
Layman description \\
    \midrule
    Perceptual confidence & Entropy of posterior over perceived states & Certainty the agent has in its inferred (perceived) state, given the data and prior knowledge (e.g. Are you sure that you are in front of the station?).\\
Utility confidence & Entropy of the probability distribution over rewards / utilities & Measures how sensitive an agent is to differences in rewards or costs (e.g. How much do I care about getting there on time verses getting a speeding ticket?)\\
Model parameter confidence & Entropy of the posterior distribution over model parameters & Certainty the agent has about its internal model parameters (e.g. How sure am I that the floor is very slippery?). \\
Control confidence&Entropy of the posterior over control signal & 
Certainty in controlling a system (e.g. How sure are you that you can control this drone in this windy condition?) \\

Decision confidence & Entropy of the posterior over discrete action options & Certainty with which someone makes a choice (e.g. How sure am I that I should turn left instead of right to get to the station?).\\[3pt]
    \bottomrule
  \end{tabular}
\end{table*}

\subsection{Robot metacognitive architecture}

The planning-monitoring-evaluation cycle from metacognition can be mapped to any cognitive skill or engineering process. For example, we can draw inspiration from the tool invention process in humans, where the cycle maps to design-test-evaluate.  Figure \ref{fig:summary}a, left and right, describes the general process in humans and how it is mapped to its robotic counterpart. One of the key components of our metacognitive architecture is the confidence evaluator. Confidence-based systems implement metacognition by adding a monitoring layer for cognitive and physical processes and evaluate both its performance and confidence. This parallels the metacognitive loop of monitoring and control in humans~\cite{Nelson1994}, but instantiates these processes as explicit computational nodes within the cognitive process cycle: an evaluator implements metacognitive monitoring (evaluating performance and confidence), while a planner/controller implements metacognitive control (adapting strategies based on those assessments). 

For concreteness but without loss of generality, we describe the robot metacognitive architecture exemplified for the tool invention process (Figure \ref{fig:summary}a, right), which consists of a loop between three components: designer (that is, planner or hypotheses generator), user and evaluator. The \textit{designer} represents the robot’s creative generative thought process, proposing new tool geometries based on current knowledge. The \textit{user} embodies the robot action system that interacts with the simulated environment to test how well the tool performs the task. Finally, the \textit{evaluator} acts as the metacognitive layer (reflection) that evaluates confidence in various key points of the decision-making process, thereby identifying environmental uncertainties allowing the agent to determine whether to trust, refine, or discard a tool design. This feedback loop allows the robot not just to iterate on physical parameters but also to think about its own thinking, adapting its design behavior based on confidence and experience. Through this metacognitive cycle, the robot evolves from a passive optimizer to an active inventor, capable of self-assessment and creative improvement in tool design and use. 

\subsection{Confidence Taxonomy and Uses}
Confidence can take different forms depending on where in the decision process it is derived (i.e. perceptual states, utility function, model parameters, control signal, and action decision). 
Table~\ref{tab:conf} summarizes these confidence channels and identifies at what point in the decision process the confidence is computed. Formally, we define confidence as the entropy of the posterior over the modeled distributions, representing the uncertainty of the system in its own inferences or choices. Incorporating this quantity into the objective function of the decision-maker (the optimization routine inside the designer in most robot problems) enables the agent to think about its own thought process, thereby endowing it with metacognitive capabilities. In effect, the agent 'thinks about its own thinking,' balancing performance rewards with robustness. Intuitively, it becomes inclined toward confident, high-reward decisions rather than risky, uncertain ones, and therefore yields robust behavior that is both adaptive and risk-aware.

We identify five uses of confidence: $i)$ \textit{as a modulator} -- when confidence in bare-hand control falls, the robot knows it must seek a new strategy,  or exploring new hypotheses, such as inventing new tools; $ii)$ \textit{as a resource allocator } -- robots, like humans, shouldn’t over-engineer when confidence is high, saving tool design only for uncertain cases, $iii)$ \textit{as a guide for control refinement} -- low confidence in predicted dynamics can motivate reshaping control through tool adjustments such as leverage, hooks, or wedges, $iv)$ \textit{as a learning signal} -- just as people learn more when unsure, low confidence can drive robots to adapt its learning rate to explore and test tool designs more aggressively, and $v)$ \textit{as a decision mechanism} -- confidence helps choose not just the best-performing tool, but the one most likely to generalize for the task under uncertainties. Confidence thus acts as a metacognitive compass, guiding robots during tool use, and triggering a tool invention when necessary.

 \section{METACOGNITION FOR TOOL INVENTION}

Tool-related cognition encompasses a wide range of fundamentally different computational problems, which we suggest to separate into three aspects: 1) designing tool candidates through optimization, 2) discovering new tools via combination of affordances of existing tools, and 3) inventing new tool structures via generative processes. Each of them requires specialized solutions, but we propose that metacognitive confidence signals can serve as a general principle to make each problem more tractable. In design, it provides continuous optimization signals; in discovery, confidence triggers model restructuring; in invention, it constrains search while guiding exploration. These three processes form a progression from constrained to increasingly open-ended search spaces. Design refines known tool concepts through parameter optimization, discovery recombines existing elements through compositional generalization, and invention generates novel structural configurations transcending existing categories. Figure \ref{fig:summary} shows the schematic of the three subproblems and how confidence contributes to their solutions. The next section will focus on providing the details of the solutions.

\subsection{Autonomous Tool Design}
Autonomous tool design refers to the process of obtaining the best tool design for a specific task given a set of environmental or physical constraints. In this section, we use previous works on tool control, selection, and design to support our proposed architecture (Figure  \ref{fig:summary}b) based on control confidence.    


\subsubsection{Tool Control, Selection and Design.}

By embedding control confidence into robotic decision making, a robot can assess how certain it is about its predicted outcomes or control strategies, much like how humans gauge confidence before acting. In \textit{tool design}, confidence can regulate exploration, driving creative adaptation when uncertainty is high, and refining precision when confidence grows. During \textit{tool selection}, confidence allows the robot to evaluate which tool best fits a task under uncertain conditions, balancing the task performance of the tool with its reliability. In \textit{tool control}, confidence adaptation acts as an internal feedback mechanism, allowing the system to dynamically adjust control gains, grip forces, or trajectories based on its control confidence. 
Control confidence can be modeled, within an approximate Bayesian inference context~\cite{Fleming2024}, as a measure of the second-order statistic of the control signal. Intuitively, it reflects the
agent’s confidence in its controller towards task completion under a given tool~\cite{meera2024confidence}. According to active inference, the precision (posterior inverse covariance matrix) of the control signal can be mathematically evaluated as the second gradient of the free energy of the agent~\cite{meera2025designing}.

Tool selection, tool control/use, and tool design are related to evaluator, user, and designer blocks, respectively. i) In \textit{tool selection}, the evaluator gauges its confidence in tool performance (for the task) to select the best tool from a toolbox. It can trigger the designer to explore alternative tools when confidence is low for all available tools. ii) In \textit{tool control}, confidence adaptation enables a robot to dynamically adjust user actions (like grip force, motion precision, or exploration rate) based on how certain it is about its current actions. This allows it to act cautiously under uncertainty and assertively when confident, much like humans refining their movements through self-assured feedback. iii) Finally, in \textit{tool design}, low confidence in performance or generalization triggers the designer to innovate by altering shapes, materials, or functional principles to create better solutions. By continuously cycling through design, use, and evaluation, robots can develop a self-reflective understanding of their own capabilities, mirroring the way humans invent new tools when existing ones fail. This metacognitive grounding enables robots not only to act intelligently but also to reason about their own competence, laying the foundation for autonomous creativity in tool invention.

\subsubsection{Advantages of using control confidence.}
Previous work has shown the advantages of using confidence in tool selection, control, and design. In the context of tool selection, the evaluator was shown to use control confidence as an early indicator of tool performance, bypassing the user block (which is often computationally expensive) from testing the tool \cite{meera2024confidence}. From a tool use perspective, the adaptation of control confidence has been shown to improve controller performance by removing any biases (priors) in the agent \cite{anil2024metacognitive}. For tool design, the use of control confidence during tool design was shown to result in tools that are more robust to environmental uncertainties during tool use \cite{meera2025designing}. Figure \ref{fig:summary}b shows the schematic for solving the tool design problem of bending a stick to form the right tool shape that can be used to pull an otherwise unreachable object in a simulated environment \cite{meera2025designing}. The designer block optimizes the tool design and generates the 3D model of the tool. The user block uses a robot arm to hold the tool and manipulate the object using a high-level robot planner and a low-level active inference controller. The evaluator monitors the task performance and evaluates the control confidence of the tool. This information is used by the designer to adapt the tool design. The use of control confidence within the evaluator contributed to the design of tools that were more robust to environmental perturbations \cite{meera2025designing}. Designing tools based on pure performance resulted in tools that fail more under uncertainties than those tools that were designed by balancing tool performance and control confidence. This highlights the importance of using control confidence to design robust tools. 

\subsection{Autonomous Tool Discovery}
Another interesting area in which our proposed architecture may be applied is for the online discovery of tools via the process of affordance combination. The concept of affordances was first introduced in psychology to describe the opportunities for action that an object offers an agent. For instance, we would consider a metal rod to have multiple affordances, such as potential for grasping, poking, and hitting as well as the possibility for being used to support or lever other objects. All of these depend uniquely on the user's own physical abilities. When it comes to tools, understanding an object's affordances and how they might be repurposed in new contexts is essential for the task of tool discovery.

Here, we apply our framework for the problem of online tool discovery in discrete planners such as partially observable Markov decision processes (POMDPs) (Figure \ref{fig:summary}c). Before an agent can discover new tools, it must first learn a world model that captures the dynamics of the tools currently available to it in its toolbox. This model captures the relationships between the agent’s actions on the objects in its environment and the resulting outcomes. Learning these tool-specific dynamics in a simulated environment allows the agent to predict how actions transform the world when using different tools. However, there may be instances in which an agent might be required to go beyond the current toolbox and come up with a novel tool for a given task. We propose that confidence can play a central role in facilitating this process. 

First, confidence could be used to identify when an impasse has been reached. In this setting, two distinct types of confidence are required. The confidence in the model parameters reflects the agent's belief in the model correctness and its understanding of the tool dynamics. 
Decision confidence in tools reflects the reliability of the predicted outcome of using a tool to achieve a goal given the information about the task at hand. Low decision confidence in tools alone can be used to prompt further learning in the environment (i.e., exploration of existing tools). In contrast, persistent low confidence in tools combined with high confidence in model parameters indicates that the agent is confident that its model is accurate, but that none of its current tools will succeed. This combination of confidence judgments can be used to trigger the designer module. 

The designer acts to prune tool-specific features in its generative model through a process of structure learning, retaining only the affordance-level representations that generalize across tools. Here, decision confidence plays a central role. It can act to guide the augmentation and pruning process, ensuring that the agent only retains representations that improve the certainty of explanatory power while discarding those that do not \cite{FRISTON2024108891}. By these means, tools become represented as structured combinations of affordances. This process could equip the agent with the ability to search over this space. By inducing an appropriate combination of affordances to solve the task, the agent can effectively produce a novel functional tool that does not exist in its current toolbox. This builds on previous work that demonstrates that appropriate representations of affordances in an agent's probabilistic generative model can enable tool innovation through compositional generalization of features \cite{collis2023understanding}.

\subsection{Autonomous Tool Invention}
In this section, we propose two use cases for confidence signals within existing tool-invention frameworks that use generative AI.
 
\subsubsection{Task-conditioned tool design under constraints.}
When it comes to ad-hoc tool design, one promising direction is the field of task-conditioned generative inverse design. This approach adopts generative architectures that can effectively capture the distribution over good solutions, which we can then sample designs from \cite{vlastelica2023diffusiongenerativeinversedesign}. We can apply scalable generative architectures such as diffusion models or generative adversarial networks to efficiently capture the statistical regularities of physically valid designs. This model learns the underlying distribution of the tool design space without regard to any specific task. Sampling from this unconditioned tool design prior therefore produces random but plausible tools, providing a foundation for further goal-directed fine-tuning when provided with information about a specific task.

We propose that our metacognitive architecture can be used to fine-tune the output of the generative model to optimize the designs of task-conditioned tools (Figure \ref{fig:summary}d). In each iteration, the designer samples candidate tool designs from the generative model. These are instantiated in the simulated world model by the user, where the agent learns to use the tool. The resulting sensory data is passed to the evaluator module, which estimates task performance and provides confidence judgments. This information then guides the designer’s constrained optimization over the tool design space. It is used to update the parameters of the generative model such that the next iteration of the sampled tools is more likely to give high-rewarding tool designs for the task at hand.

As this iterative fine-tuning is computationally expensive, we propose the use of confidence to improve efficiency in this process. Inspired by previous work on feedback-efficient fine-tuning in the context of task-conditioned design, we identify two opportunities to use confidence-aware mechanisms. Both are related to the problem of efficient exploration of a high-dimensional design space to produce successful designs \cite{uehara2024feedbackefficientonlinefinetuning}. First of all, we could use model parameter confidence to identify areas of uncertainty in the downstream reward, and therefore areas of the design space that warrant exploration. This can be achieved by calculating the confidence of the task performance block in the evaluator. Low confidence in the predicted reward of a given design is assigned a higher value, providing intrinsic motivation to resolve uncertainty in the design space. This active learning helps encourage efficient coverage of the design space in order to hone in on good solutions.

Secondly, we could use confidence to regularize the update of the generative model. During fine-tuning, it is important to make sure that the designer adheres to its prior knowledge over physically valid designs and does not stray too far out-of-distribution (i.e. does not greedily optimize for reward). Here, confidence can serve as a signal that can be used to dynamically adjust the penalty for deviating too far from the prior. This effectively modulates a trust region (which describes the safe set of changes that can be made to the generative model). When the evaluator expresses high confidence in its task performance, the model can adaptively relax this prior constraint and exploit promising regions of the design space. In contrast, when confidence is low, the penalty can be increased in order to keep updates to the generative model close to well-understood regions of the design space. This process can be understood as the use of confidence to balance the need for efficient fine-tuning with the capacity for innovation. This flexible mechanism invests in the idea that a designer should allocate design effort adaptively and explore only when appropriate.

\subsubsection{Confidence-aware generative inverse design of tools.}
We propose a confidence-based mechanism to guide generative design processes. This mechanism could help determine the candidate designs to pursue or discard, thus aiding in optimization. Within our framework, a generative \textit{designer} produces candidate tool geometries, an \textit{evaluator} estimates each design’s likelihood of success for a given task, and a \textit{user module} (e.g., a simulation or a real-world test) provides contextual feedback.

The evaluator can assess confidence by integrating multiple signals, particularly those related to uncertainty. (i) Epistemic uncertainty captures the model’s lack of knowledge about the design space. High epistemic uncertainty arises when the system encounters unfamiliar or poorly understood designs. Reducing this type of uncertainty usually requires more data or exploration \cite{hullermeier2021aleatoric}. (ii) Aleatoric uncertainty refers to inherent ambiguity in the task or environment cases where multiple valid designs could exist or where noise is unavoidable. This uncertainty cannot be reduced simply by collecting more data \cite{hullermeier2021aleatoric}. Basic validity checks can include lightweight constraints to ensure that the design is feasible or syntactically correct before further computations. The evaluator could then combine these sources of information to produce a single confidence score (decision confidence), interpreted as the estimated probability that the design will succeed. This aligns with Bayesian interpretations of confidence as a second-order belief about correctness \cite{Meyniel2015}. In practical implementations, this confidence score can be calibrated using techniques such as temperature scaling to ensure its reliability \cite{guo2017calibration}.

Example workflow (Figure~\ref{fig:summary}(e) visually describes it): As a generative model iteratively constructs a design, the evaluator could estimate epistemic and aleatoric uncertainties at key checkpoints. Based on the evolving confidence score:
(i) If confidence is low, the system might take more conservative steps, switch strategies, or discard the current design early. (ii) If confidence is high, it might proceed more aggressively or halt early to conserve resources. After generating a batch of candidates, the designs that fail basic checks are removed. The remaining valid designs are ranked by confidence, and only the most promising ones are subjected to further refinement or optimization. This confidence-guided filtering could lead to faster convergence, reduced computational costs, and more robust outcomes, especially in domains where generative search spaces are large and uncertain.

\section{DIRECTIONS}

When performance monitoring and confidence evaluation are explicit in the cognitive architecture, the system can $i)$ propagate these signals to other computational modules, (ii) communicate with humans, and (iii) share between agents. That is, the robot can express its confidence in its cognitive and physical actions. This improves reliability in decision making while fostering greater awareness and transparency in robotic behavior. In this section, we propose complementary future research directions for using confidence in decision making.

\subsection{Confidence for physical intelligence}
Metacognitive mechanisms, particularly confidence signals, can be deployed in different computational challenges, from low-level sensory processes to high-level strategic decision making \cite{Fleming2024}. 
 However, most studies examine human metacognition in well-defined problem spaces, focus on confidence modulating existing operations rather than generating novel structures, and study these functions in isolation rather than their coordination across compositional, optimization, and generative regimes. How humans use these metacognitive mechanisms in uncertain scenarios involving physical interaction, and particularly, their role in creative processes, such as tool invention, remains an open question. 

\subsection{Confidence for autonomous design and manufacturing pipelines}
Autonomous design–manufacturing pipelines link generative design, simulation (often through digital twins), and automated fabrication, especially additive manufacturing (3D printing), to turn requirements into candidate parts and tools with few manual steps. Bringing confidence into the design stage helps prioritize what to simulate and prototype, cuts trial-and-error cycles, shortens design–build–test cycles, accelerates innovation in machine design, and supports co-design between engineers and robots. The same confidence signal can be used to allocate resources and build time to the most promising designs and can route low-confidence plans to revision before an expensive printing. The result is a higher deployment efficiency and improved sustainability (fewer reprints and aborted builds, better material use, and faster, more predictable schedules) in additive manufacturing technologies and large-scale construction projects.

\subsection{Trustworthy robot decision making}
Future robotic systems will increasingly rely on confidence as a cornerstone of decision making. This can help them act safely, transparently, and adaptively in complex environments. In human–robot collaboration, confidence reporting can allow robots to communicate their uncertainty for trustworthy and shared decision-making. In autonomous driving, confidence-aware control could help vehicles navigate uncertainty by refining the controller, seeking more information from sensors, or transferring control when confidence drops. Similarly, in safety-critical applications, such as surgical robotics, confidence becomes a measure of risk awareness, allowing systems to dynamically balance performance and caution (risk aversion). By quantifying and communicating how sure they are about their perceptions and actions, future robots will be able to make better decisions and become more trustworthy and interpretable in complex environments.

\subsection{Robot Metacognition for Collective Innovation}
Metacognition plays an important role in cultural evolution \cite{Dunstone2018}. Tools have historically evolved through a cumulative culture, and technologies have developed through incremental modifications across generations. Autonomous robots with confidence-guided tool invention could complement human tool-making practices. Rather than replacing human innovation, such systems could propose novel design alternatives, test them systematically, and communicate uncertainty throughout, thereby enabling a qualitatively different mode of cumulative culture through effective human-robot collaboration.

\subsection{Robots with improved awareness}


As a key facet of a robot's metacognitive ability, confidence serves as a cornerstone for developing the full spectrum of synthetic awareness mechanisms that significantly enhance its problem-solving capabilities. This internal evaluation also allows for self-explanation and transparency, enabling robots to communicate their decisions and report their confidence levels, a crucial step in fostering trustworthy autonomous operation and human-robot collaboration. Providing explanations for their actions has been shown to increase user satisfaction and trust when working with collaborative robots. In multirobot systems, the sharing of metacognitive signals can lead to a form of collective awareness, where one robot's confidence can inform and influence the decisions of its counterparts. 

Ultimately, embedding a confidence-based metacognitive loop could transform robots from simple reactive agents into self-evaluating, self-improving designers. 
Such robots would be capable of acting with awareness, reasoning about their decisions, and inventing new solutions when their confidence is low. 


\section{ACKNOWLEDGMENTS}
This research is supported by the MetaTool project (Grant agreement 101070940) under the European Innovation Council (EIC) pathfinder program.  PFK and PC are funded by the EIC via the UKRI Horizon Europe Guarantee scheme as part of the MetaTool project.



\normalsize
\bibliography{references}


\end{document}